\documentclass[10pt,twocolumn,letterpaper]{article}

\usepackage{cvpr}
\usepackage{times}
\usepackage{epsfig}
\usepackage{graphicx}
\usepackage{amsmath}
\usepackage{amssymb}
\usepackage{multirow}
\usepackage[pagebackref=true,breaklinks=true,letterpaper=true,colorlinks,bookmarks=false]{hyperref}

\cvprfinalcopy % *** Uncomment this line for the final submission

\def\cvprPaperID{2717} % *** Enter the CVPR Paper ID here

\ifcvprfinal\fi
\begin{document}

%%%%%%%%% TITLE
\title{All You Need is Beyond a Good Init: Exploring Better Solution
 for Training Extremely Deep Convolutional Neural Networks with
 Orthonormality and Modulation}

\author{Di Xie\\
{\tt\small xiedi@hikvision.com}
\and
Jiang Xiong\\
{\tt\small xiongjiang@hikvision.com}\\
Hikvision Research Institute\\
Hangzhou, China
\and
Shiliang Pu\\
{\tt\small pushiliang@hikvision.com}
}

% For a paper whose authors are all at the same institution,
% omit the following lines up until the closing ``}''.
% Additional authors and addresses can be added with ``\and'',
% just like the second author.
% To save space, use either the email address or home page, not both
%\and
%Jiang Xiong\\
%Hikvision Research Institute\\
%\\
%{\tt\small xiongjiang@hikvision.org}
%}

\maketitle
%\thispagestyle{empty}

%%%%%%%%% ABSTRACT
\begin{abstract}
   Deep neural network is difficult to train and this predicament becomes
   worse as the depth increases. The essence of this problem exists in the
   magnitude of backpropagated errors that will result in gradient vanishing
   or exploding phenomenon. We show that a variant of regularizer which
   utilizes orthonormality among different filter banks can alleviate this
   problem. Moreover, we design a backward error modulation mechanism based
   on the quasi-isometry assumption between two consecutive parametric layers.
   Equipped with these two ingredients, we propose several novel optimization
   solutions that can be utilized for training a specific-structured (repetitively
   triple modules of Conv-BN-ReLU) extremely
   deep convolutional neural network (CNN) \emph{WITHOUT} any shortcuts/
   identity mappings from scratch.
   Experiments show that our proposed solutions can achieve distinct improvements for a
   44-layer and a 110-layer plain networks
   on both the CIFAR-10 and ImageNet datasets. Moreover, we can successfully train
   plain CNNs to match the performance of the residual counterparts.

   Besides, we propose new principles for designing network structure from the insights
   evoked by orthonormality. Combined with residual structure, we achieve
   comparative performance on the ImageNet dataset.
\end{abstract}

%%%%%%%%% BODY TEXT
\section{Introduction}

Deep convolutional neural networks have improved performance across a wider
variety of computer vision tasks, especially for image
classification~\cite{Krizhevsky2012ImageNet,Simonyan2014Very,Szegedy2015Going,
Sermanet2013OverFeat,Zeiler2013Visualizing},
object detection~\cite{yang2016craft,Ren2016Faster,ShrivastavaGG16}
and segmentation~\cite{long2015fully,chen2016deeplab,pinheiro2016learning}.
Much of this improvement
should give the credit to gradually deeper network architectures. In just
four years, the layer number of networks escalates from several to hundreds,
which learns more abstract and expressive representations from large amount
of data, \eg~\cite{ILSVRC15}. Simply stacking more layers onto current
architectures is not a reasonable solution, which incurs vanishing/exploding
gradients~\cite{Bengio,Glorot2010Understanding}. To handle the relatively
shallower networks, a
variety of initialization and normalization methodologies are
proposed~\cite{Glorot2010Understanding,Saxe,He2015Delving,Sussillo2015Random,
Kr2015Data,Mishkin2015All,Ioffe2015Batch,Arpit2016Normalization}, while deep
residual learning~\cite{He2015Residual} is utilized to deal with
extremely deep ones.

Though other works, \eg~\cite{Srivastava2015Deep,Srivastava2015Highway}, have
also announced that they can train an extremely deep network with improved
performance, deep residual network~\cite{He2015Residual} is still the best and
most practical solution for dealing with the degradation of training accuracy
as depth increases. However, it is substantial that residual networks are
exponential ensembles of relatively shallow ones (usually only 10-34 layers
deep), as an interpretation by Veit \etal~\cite{Veit2015Exp}, it avoids the
vanishing/exploding gradient problem instead of resolving it directly.
Intrinsically, the performance gain of networks is determined by its
multiplicity, not the depth. So how to train an ultra-deep network is still an
open research question with which few works concern. Most researches still focus
on designing more complicated structures based on residual block and its
variants~\cite{Larsson2016FractalNet,Zagoruyko2016Wide}.
Anyway, dose there exist an applicable methodology that can be used for training
a genuinely deep network?

\begin{figure}[t]
\begin{center}
%\fbox{\rule{0pt}{2in} \rule{0.9\linewidth}{0pt}}
   \includegraphics[width=0.4\linewidth]{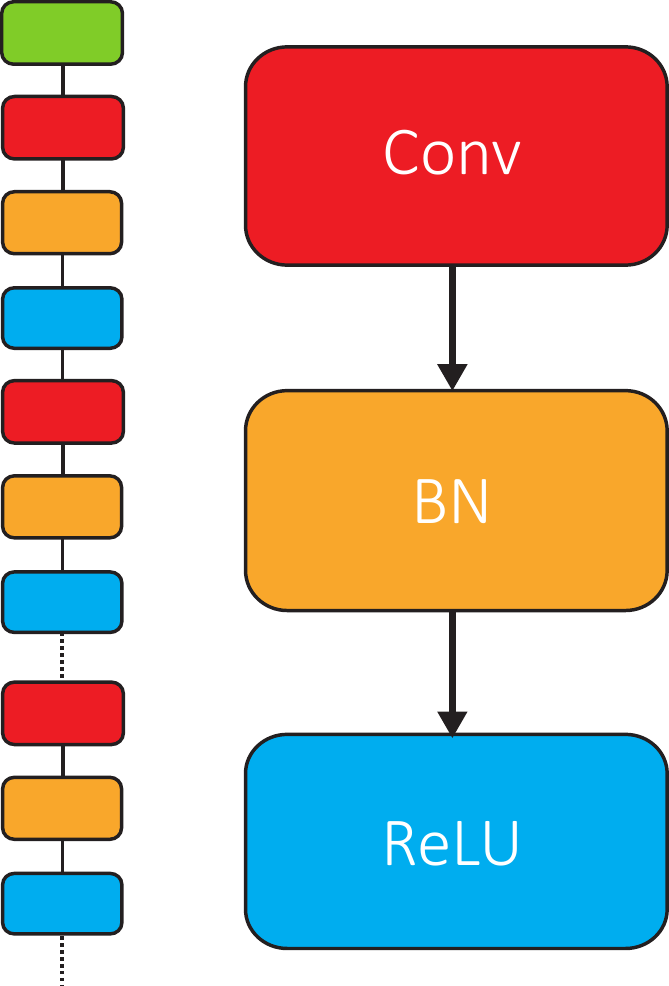}
\end{center}
   \caption{Diagram of the plain CNN network architecture (left) and repetitive
   triple-layer module (right) in this paper. Green box is for input data,
   Red color ones denotes
   parametric layers (convolutional or fully connected), yellow
   represents batch normalization layers and
   blue means activation layers. Actually, this structure is similar
   with the plain CNN designed by He \etal~\cite{He2015Residual}.}
\label{fig:networkmodule}
\end{figure}

In this paper, we try to find a direct feasible solution to answer above question.
We think batch normalization (BN)~\cite{Ioffe2015Batch} is necessary to ensure
the propagation stability in the forward pass in ultra-deep networks and the
key of learning availability exists in the backward pass which propagates
errors with a top-down way. We constrain the network's structure to repetitive
modules consisted by Convolution, BN and ReLU~\cite{Nair2010ReLU} layers
(Fig.~\ref{fig:networkmodule}) and analyze the Jacobian of the output with respect to
the input between consecutive modules. We show that BN cannot guarantee the magnitude
of errors to be stable in the backward pass and this amplification/attenuation effect to
signal will accumulate layer-wisely which results in gradients exploding/vanishing.
From the view of norm-preserving, we find that keeping the orthonormality between
filter banks within a layer during learning process is a sufficient and necessary
condition to ensure the stability of backward errors. While this condition cannot
be satisfied in nonlinear networks equipped with BN, this orthonormal constrain can
mitigate backward signal's attenuation and we prove it by experiments. An orthonormal
regularizer is introduced to replace traditional weight decay regularization~\cite{Girosi95Reg}.
Experiments show that there is $3\%\thicksim4\%$ gains for a 44-layer network
on CIFAR-10.

However, as depth increases, \eg deeper than 100 layers, the non-orthogonal
impact induced by BN, ReLU and gradients updating accumulates, which breaks
the dynamic isometry~\cite{Saxe} and makes learning unavailable. To neutralize
this impact, we design a modulation mechanism based on the quasi-isometry
assumption between two consecutive parametric layers. We show the
quasi-isometry property with both mathematical analysis and experiments.
With the modulation, a global scale factor can be applied on the magnitude
of errors a little unscrupulously during the backward pass in a layer-wise fashion.
Combined with orthonormality, experiments show that a plain CNN shown
in Fig.~\ref{fig:networkmodule} can be trained relatively well and match the
performance of its residual counterpart.

The contributions of this paper are summarized as follows. 1) We demonstrate the
necessity of applying BN and explain the potential
reason which results in degradation problem in optimizing deep CNNs;
2) A concise methodology equipped with orthonormality and modulation
is proposed to provide more insights to understand learning dynamics of CNNs;
3) Experiments and analysis exhibit interesting phenomenons and promising
research directions.

%-------------------------------------------------------------------------

\section{Related Work}
\textbf{Initialization in Neural Networks.} As depth increases, Gaussian
initialization cannot suffice to train a network from scratch~\cite{Simonyan2014Very}.
The two most prevalent works are proposed by Glorot \& Bengio~\cite{Glorot2010Understanding}
and He \etal~\cite{He2015Delving} respectively. The core idea of their
works is to keep the unit variance of each layer's output.
Sussillo \& Abbott~\cite{Sussillo2015Random} propose a novel random walk initialization
and mainly focus on adjusting the so-called scalar factor $g$ to make the ratio of
input/output error to be constant around $1$.
Kr\"ahenb\"uhl \etal~\cite{Kr2015Data} introduce data-dependent initialization to ensure all
layers training at an equal rate.

Orthogonality is also in consideration. Saxe \etal~\cite{Ganguli2013Learning,Saxe} analyse
the dynamics of learning in linear deep neural networks. They find that the convergence rate
of random orthogonal initialization of weights is equivalent to unsupervised pre-training,
which are both superior to random Gaussian initialization. LSUV initialization
method~\cite{Mishkin2015All} is proposed which not only takes advantage of orthonormality but
also makes use of the unit-variance of each layer's output.

In our opinion, a well-behaved initialization is not enough to resist the variation as
learning progresses, which is to say, to have a good initial condition (\eg isometry) cannot
ensure the preferred condition to keep unchanged all the time, especially in extremely deep
networks. This argument forms the basic idea that motivates us to explore the solutions for
genuinely deep networks.

\textbf{Signal Propagation Normalization.}
Normalization is a common and ubiquitous technique in machine learning community. The
whitening and decorrelation of input data brings benefits to both deep learning and other
machine learning algorithms, which helps speeding up the training process~\cite{Lecun2000Efficient}.
Batch normalization~\cite{Ioffe2015Batch} generalize this idea to ensure
each layer's output to be identical distributions which reduce the internal
covariate shift. Weight normalization~\cite{Salimans2016Weight}
is inspired by BN by decoupling the norm of the weight vector from its
direction while introducing independencies between the examples in a minibatch.
To overcome the disadvantage of BN that dependent on minibatch size, layer
normalization~\cite{Ba2016Layer} is proposed to solve the normalization problem
for recurrent neural networks. But this method cannot be applied to CNN, as
the assumption violates the statistics of the hidden layers. For more
applicable in CNN, Arpit~\etal introduce normalization
propagation~\cite{Arpit2016Normalization} to reduce the internal covariate shift for
convolutional layers
and even rectified linear units. The idea of normalization each layers'
activations is promising, but a little idealistic in practice. Since the
incoherence prior of weight matrix is actually not true in the initialization
phase and even worsen in iterations, the normalized magnitude of each layer's
activations cannot be guaranteed in an extremely deep network. In our
implementation, it even cannot prevent the exploding activations' magnitude
just after initialization.

\textbf{Signal Modulation.}
Few work is done in this field explicitly, but implicitly integrated the
idea of modulation. In a broad sense, modulation can be viewed as a persistent
process of the combination of normalization and other methodology to keep
the magnitude of a variety of signals steady at learning. With this
understanding, we can summarize all the methods above with a unified framework,
\eg batch normalization~\cite{Ioffe2015Batch} for activation modulation,
weight normalization~\cite{Salimans2016Weight} for parameter modulation, \etc.

%-------------------------------------------------------------------------
\section{Methodology}
%-------------------------------------------------------------------------
\subsection{Why is BN a requisite?}\label{sec:BN}
Since the complexity dynamics of learning in nonlinear neural
networks~\cite{Saxe}, even a proven mathematical theory cannot guarantee
that a variety of signals keeping isometrical at the same time in practice
applications. Depth itself results in the ``butterfly effect" with exponential
diffusion while nonlinear gives rise to indefiniteness and randomness. Recently
proposed methods~\cite{Arpit2016Normalization,Sussillo2015Random,Kr2015Data}
which utilize isometry fail to keep the steady propagation of signals in
over-100-layer networks. These methods try to stabilize the magnitude of signals
from one direction (forward/backward) as a substituted way to control the signals
in both directions. However, since the complexity variations of signals, it
is impossible to have conditions held on both ways with just one modulation
method.

An alternative option is to simplify this problem to constrain the magnitude
of signals in either direction, which we can pay the whole attention to another
direction\footnote{For a specified weight that connected $i$th neuron in $l$th
layer and $k$th neuron in $(l+1)$th layer, $w^{(l)}_{ij}$, its gradient can be
computed as $\nabla~w^{(l)}_{ij}=a^{(l)}_{i}\times\delta^{(l+1)}_{j}$.
If the two variables are independent from each other, then the magnitude of
gradient can be directly related with just one factor~(activation/error).}.
Batch normalization is an existed solution that satisfies our
requirement. It does normalization in the forward pass to reduce internal
covariate shift with a layer-wise way\footnote{Methods modulate
signals without a layer-wise manner, \eg~\cite{Arpit2016Normalization},
will accumulate the indefiniteness with a superlinear way and finally the
propagated signals will be out of control.}, which, in our opinion, make us to
focus all the analyses on the opposite direction.

From~\cite{Ioffe2015Batch}, during the backpropagation of the gradient of loss
$\ell$ through BN, we can formulate errors between adjacent layers as follow:
\begin{equation}
\frac{\partial~\ell}{\partial~x_{i}}=\frac{1}{\sqrt{\sigma_{B}^{2}+\epsilon}}
(\delta_{i}-\mu_{\delta}-\frac{\hat{x}_{i}}{m}\sum_{j=1}^{m}\delta_{j}\hat{x}_{j})
\label{equation1}
\end{equation}
where $x_{i}$ is $i$th sample in a mini-batch (we omit activation index for
simplicity), so
$\frac{\partial~\ell}{\partial~x_{i}}$ denotes output error.
$\delta_{i}=\frac{\partial~\ell}{\partial~y_{i}}\cdot\gamma$ where
$\frac{\partial~\ell}{\partial~y_{i}}$ is the input error and $\gamma$ is
scale parameter of BN. $\mu_{\delta}=\frac{1}{m}\sum_{i=1}^{m}\delta_{i}$
is mean of scaled input errors, where $m$ denotes mini-batch's size.
$\hat{x}_{i}=\frac{x_{i}-\mu_{B}}{\sqrt{\sigma_{B}^{2}+\epsilon}}$ is the
corresponding normalized activation.

Equation~\ref{equation1} represents a kind of ``pseudo-normalization"
transformation for error signals $\delta_{i}$ compared with its forward
operation. If the mean of distribution of input error $\delta_{i}$ is
zero and symmetric, we can infer that the mean of distribution of output
error is approximately zero. It centralizes the errors and the last term $\frac{\hat{x}_{i}}{m}\sum_{j=1}^{m}\delta_{j}\hat{x}_{j}$ will bias
the distribution but these biases may be cancelled out from each other
owing to the normalized coefficient $\hat{x}_{i}$ which is normal distribution.
Besides, errors are normalized with a mismatched
variance. This type of transformation will change error signal's original
distribution with a layer-wise way since the second order moment of each layer's
output errors loses its isometry progressively.
However, this
phenomenon can be ignored when we only consider a pair of consecutive layers.
In a sense, we can think the backward propagated errors are also normalized
as well as its forward pass, which is why we apply ``Conv-BN-ReLU" triple
instead of ``Conv-ReLU-BN"\footnote{Another reason is that placing ReLU after
BN guarantees approximately $50\%$ activations to be nonzero, while the ratio
may be unstable if putting it after convolution operation.}.

The biased distribution effect will accumulated as depth increases and distort
input signals' original distribution, which is one of several reasons that make
training extreme deep neural network difficult. In next section we try to solve
the problem to some extent.
%-------------------------------------------------------------------------
\subsection{Orthonormality}
Norm-preserving resides in the core idea of this section. A vector
$\textbf{x}\in\Re^{d_{\textbf{x}}}$ is mapped by a linear transformation
$\textbf{W}\in\Re^{d_{\textbf{x}}\times~d_{\textbf{y}}}$ to another vector
$\textbf{y}\in\Re^{d_{\textbf{y}}}$, say, $\textbf{y}=\textbf{W}^{T}\textbf{x}$.
If $\|\textbf{y}\|=\|\textbf{x}\|$, then we call this transformation
norm-preserving. Obviously, orthonormality, not the normalization proposed
by~\cite{Arpit2016Normalization} alone, is both sufficient and necessary
for holding this equation, since
\begin{equation}
\|\textbf{y}\|=\sqrt{\textbf{y}^{T}\textbf{y}}=\sqrt{\textbf{x}^{T}\textbf{W}
\textbf{W}^{T}\textbf{x}}=\sqrt{\textbf{x}^{T}\textbf{x}}=\|\textbf{x}\|~iff.~
\textbf{W}\textbf{W}^{T}=\textbf{I}
\label{equation2}
\end{equation}

Given the precondition that signals in forward pass are definitely normalized,
here we can analyse the magnitude variation of errors only in backward pass.
To keep the gradient with respect to the input of previous layer norm-preserving,
it is straightforward to conclude that we would better maintain orthonormality among
columns\footnote{Beware of the direction, which results in the exchange of notations
in equation~\ref{equation2}. So the rows and columns of the matrix are also exchanged.}
of a weight matrix in a specific layer during learning process rather
than at initialization according to Eq.~\ref{equation2}, which equivalently makes
the Jacobian to be ideally dynamical isometry~\cite{Saxe}. Obviously in CNN this
property
cannot be ensured because of 1) the gradient update which makes the correlation
among different columns of weights stronger as learning proceeding; 2) nonlinear
operations, such as BN and ReLU, which destroy the orthonormality. However, we
think it is reasonable to force the learned parameters to be conformed with the
orthogonal group as possible, which can alleviate vanishing/exploding phenomenon of
the magnitude of errors and the signal distortion after accumulated nonlinear
transformation. The rationality of these statements and hypotheses has been proved
by experiments.

To adapt the orthonormality for convolutional operations, we generalize the orthogonal
expression with a direct modification. Let $\tilde{\textbf{W}}_{l}\in\Re^{W\times~H\times~C\times~M}$
denote a set of convolution kernels in $l$th layer, where $W$, $H$, $C$, $M$
are width, height,
input channel number and output channel number, respectively. We replace original
weight decay regularizer with the orthonormal regularizer:
\begin{equation}
\frac{\lambda}{2}\sum_{i=1}^{D}\|\textbf{W}_{l}^{T}\textbf{W}_{l}-\textbf{I}\|_{F}^{2}
\label{equation3}
\end{equation}
where $\lambda$ is the regularization coefficient as weight decay,
$D$ is total number of convolutional layers and/or fully connected layers,
$\textbf{I}$ is the identity matrix and $\textbf{W}_{l}\in\Re^{f_{in}\times~f_{out}}$
where $f_{in}=W\times~H\times~C$ and $f_{out}=M$. $\|\cdot\|_{F}$ represents the Frobenius norm.
In other words, equation~\ref{equation3}
constraints orthogonality among filters in one layer, which makes the learned features
have minimum correlation with each other, thus implicitly reduce the redundancy and
enhance the diversity among the filters, especially those from the lower
layers~\cite{Shang2016Understanding}.

Besides, orthonormality constraints provide alternative solution other than $L2$
regularization to the exploration of weight space in learning process. It provides
more probabilities by limiting set of parameters in an orthogonal space instead of
inside a hypersphere.

\subsection{Modulation}
The dynamical isometry of signal propagation in neural networks has been mentioned
and underlined several times~\cite{Arpit2016Normalization,Saxe,Ioffe2015Batch}, and
it amounts to maintain the singular values of Jacobian, say
$\textbf{J}=\frac{\partial\textbf{y}}{\partial\textbf{x}}$, to be around $1$. In this
section, we will analyze the variation of singular values of Jacobian through different
types of layers in detail. We omit the layer index and bias term for simplicity
and clarity.

For linear case, we have $\textbf{y}=\textbf{W}^{T}\textbf{x}$, which shows that
having dynamical isometry is equivalent to keep orthogonality since
$\textbf{J}=\textbf{W}^{T}$ and $\textbf{J}\textbf{J}^{T}=\textbf{W}^{T}\textbf{W}$.

Next let us consider the activations after normalization transformation,
$\textbf{y}=\textrm{BN}_{\gamma,\beta}(\textbf{W}^{T}\textbf{x})$, which we borrow
the notation from~\cite{Ioffe2015Batch}. Given the assumption that input dimension
equals output dimension and both are $d$-dimension vectors, the Jacobian is
\begin{equation}
\textbf{J}=\left[
\begin{array}{cccc}
\textbf{J}_{11} & \textbf{0} & \cdots & \textbf{0}\\
\textbf{0} & \textbf{J}_{22} & \cdots & \textbf{0}\\
\vdots & \vdots & \ddots & \vdots\\
\textbf{0} & \textbf{0} & \cdots & \textbf{J}_{dd}
\end{array}
\right]_{md\times md}
\label{equation8}
\end{equation}
where each $\textbf{J}_{kk}$ is a $m\times m$ square matrix, that is
\begin{equation}
\textbf{J}_{kk}=\left[
\begin{array}{cccc}
\frac{\partial y_{1}^{(k)}}{\partial x_{1}^{(k)}} & \frac{\partial y_{1}^{(k)}}
{\partial x_{2}^{(k)}} &
\cdots & \frac{\partial y_{1}^{(k)}}{\partial x_{m}^{(k)}}\\
\frac{\partial y_{2}^{(k)}}{\partial x_{1}^{(k)}} & \frac{\partial y_{2}^{(k)}}
{\partial x_{2}^{(k)}} &
\cdots & \frac{\partial y_{2}^{(k)}}{\partial x_{m}^{(k)}}\\
\vdots & \vdots & \ddots & \vdots\\
\frac{\partial y_{m}^{(k)}}{\partial x_{1}^{(k)}} & \frac{\partial y_{m}^{(k)}}
{\partial x_{2}^{(k)}} &
\cdots & \frac{\partial y_{m}^{(k)}}{\partial x_{m}^{(k)}}
\end{array}
\right]
\label{equation4}
\end{equation}
Here $\frac{\partial y_{i}^{(k)}}{\partial x_{j}^{(k)}}$ denotes partial derivative
of output of $i$th sample with respect to $j$th sample in $k$th component. The
Jacobian of BN has its speciality that its partial derivatives are not only
related with components of activations, but also with samples in one mini-batch.
Because of each component $k$ of activations is transformed independently by BN,
$\textbf{J}$ can be expressed with a blocked diagonal matrix as Eq.~\ref{equation8}.
Again since the independence among activations, we can analyse just one of $d$
sub-Jacobians, \eg~$\textbf{J}_{kk}$.

From equation~\ref{equation1} we can get the entries of $\textbf{J}_{kk}$, which is
\begin{equation}
\frac{\partial y_{j}}{\partial x_{i}}=\rho\left[\Delta(i=j)-\frac{1+\hat{x}_{i}\hat{x}_{j}}{m}\right]
\label{equation5}
\end{equation}
where $\rho=\frac{\gamma}{\sqrt{\sigma_{B}^{2}+\epsilon}}$ and
$\Delta(\cdot)$ is the indicator operator. Here we still omit index
$k$ since dropping it brings no ambiguity.

Eq.~\ref{equation5} concludes obviously that
$\textbf{J}\textbf{J}^{T}\neq\textbf{I}$. So the orthonormality is not held after
BN operation. Now the correlation among
columns of $\textbf{W}$ is directly impacted by normalized activations, while the
corresponding weights determine these activations in turn, which results in a complicated
situation. Fortunately, we can deduce the preferred equation according to subadditivity
of matrix rank~\cite{Banerjee2014Linear}, which is
\begin{equation}
\textbf{J}=\textbf{P}^{T}
\rho\left[
\begin{array}{ccccc}
1-\frac{\lambda_{1}}{m} & 0 & 0 & \cdots & 0\\
0 & 1-\frac{\lambda_{2}}{m} & 0 & \cdots & 0\\
0 & 0 & 1 & \cdots & 0\\
\vdots & \vdots & \vdots & \ddots & \vdots\\
0 & 0 & 0 & \cdots & 1
\end{array}
\right]_{m\times m}
\textbf{P}
\label{equation7}
\end{equation}
where $\textbf{P}$ is the matrix consists of eigenvectors of $\textbf{J}$. $\lambda_{1}$
and $\lambda_{2}$ are two nonzero eigenvalues of $\textbf{U}$, say
$U_{ij}=1+\hat{x}_{i}\hat{x}_{j},~i=1\cdots m,~j=1\cdots m$.

Eq.~\ref{equation7} shows us that $\textbf{J}\textbf{J}^{T}\approx\rho^{2}\textbf{I}$
\footnote{The Jacobian after ReLU is amount to multiply a scalar with
$\textbf{J}$~\cite{Arpit2016Normalization}, which we can merge it into $\rho$ instead.}.
The approximation comes from first two diagonal entries in Eq.~\ref{equation7} which may be
close to zero. We think it is one of reasons that violate the perfect dynamic isometry and
result in the degradation problem with
this kind of non-full rank. Since value of $\rho$ is determined by $\gamma$ and $\sigma_{B}$,
it is bounded as long as these two variables keep stable during the learning process, which
achieves the so-called quasi-isometry~\cite{Collins1998Combinatorial}.

Notice that $\rho$ changes with $\gamma$ and $\sigma_{B}$ while $\gamma$ and $\sigma_{B}$ will
change in every iteration. Based on the observation, we propose the scale factor $\rho$ should
be adjusted dynamically instead of fixing it
like~\cite{Arpit2016Normalization,Sussillo2015Random,Saxe}. According to~\cite{Saxe}, when the
nonlinearity is odd, so that the mean activity in each layer is approximately $0$, neural
population variance, or second order moment of output errors, can capture these dynamical properties
quantitatively. ReLU nonlinearity is not satisfied but owing to the pseudo-normalization we can
regard the errors propagated backwardly through BN as having zero mean, which makes the second
order moment statistics reasonable.

%------------------------------------------------------------------------
\section{Implementation Details}
We insist to keep the orthonormality throughout the training process, so
we implement this constraint both at initialization and in regularization.
For a convolution parameter $\textbf{W}_{l}\in\Re^{f_{in}\times~f_{out}}$
of $l$th layer, we initialize subset of $\textbf{W}$, say $f_{in}$-dimension
vectors, on the first output channel. Then Gram-Schmidt process is applied
to sequentially generate next orthogonal vectors channel by channel.
Mathematically, generating $n$ orthogonal vectors in $d$-dimension space
which satisfies $n>d$ is ill-posed and, hence, impossible. So one solution
is to avoid the fan-ins and fan-outs of kernels violating the principle,
say $f_{in}\geq f_{out}$, in designing structures of networks; another
candidate is group-wise orthogonalization proposed by us. If
$f_{in}<f_{out}$, we divide the vectors into
$\frac{f_{out}}{f_{in}}+f_{out}\textrm{mod}f_{in}$ groups,
orthogonalization is implemented within each group independently. We
do not encourage the hybrid utilization of $L2$ regularization for those
parameters of $f_{in}<f_{out}$ and orthonormal regularization for those
of $f_{in}\geq f_{out}$. Forcing parameters to retract into inconsistent
manifolds may cause convergence problems. Details can be referred in
experiments.

For signal modulation, we compute the second order moment statistics of output
errors between consecutive parametric layers (convolutional layer in our
case) in each iteration. The scale factor $\rho$ is defined as the square
root of ratio of second order moment of higher layer, say $q^{l+1}$, to that,
say $q^{l}$, of lower layer. However, if we modulate all the layers as
long as $q^{l+1}\neq q^{l}$, then the magnitude of propagated signal will
tend to be identical with the input error signal, which probably eliminate
the variety encoded in the error signal. So we make a trade-off that the
modulation only happens when the magnitudes of propagated signals of
consecutive layers mismatch. Experiments show that it is a relatively
reasonable and non-extreme modulation mechanism which has a capability of
maintaining magnitude constancy for error signals.

\section{Experiments}
First of all, we must demonstrate that the core idea of this paper is to
show that the proposed methods can be used to train extremely deep and
plain CNNs and improve the performance drastically compared against
prevalent stochastic gradient descent (SGD) with $L2$ regularization
rather than achieving state-of-the-art performance in a certain dataset
by all manner of means. Moreover, we try to show that the degradation
problem of training a plain network reported
in~\cite{He2015Residual,He2014Convolutional,Srivastava2015Highway} can be
partially solved by our methods.
\subsection{Datasets and Protocols}
Two representative datasets, CIFAR-10~\cite{Krizhevsky2012Learning} and
ImageNet~\cite{ILSVRC15}, are used in our experiments.

\textbf{CIFAR-10.}
CIFAR-10 consists of
$60,000$ $32\times32$ real world color images in $10$ classes split into
$50,000$ train and $10,000$ test images. All present experiments are trained
on the training set and evaluated on the test set. Top-1 accuracy is evaluated.

\textbf{ImageNet 2012 classification.}
For large-scale dataset, ImageNet 2012 classification
dataset is used in our experiments. It consists of $1000$ classes and there
are $1.28$ million training images and $50$k validation images. Both top-1
and top-5 error rates are evaluated.

\textbf{Protocol of CIFAR-10.}
To demonstrate that our proposed method can partially solve the degradation
problem and show that the gap between deeper plain network and the shallower
one can be shrunk or even removed, we aim to have fair comparison
%\footnote{
%Here ``fair comparison" means the baseline we tend to compare is not from the
%authors' results in their paper, while is implemented by ourselves. So having
%a minor different architecture does not matter.}
with the
plain network in~\cite{He2015Residual}. So we directly adopt their proposed
architectures with minor modifications for both plain networks and residual
networks. Specifically, the network inputs are $32\times32$ images, with the
per-pixel mean subtracted and standard deviation divided. The first layer is
$3\times3$ convolution and then following a stack of $6n$ $3\times3$ convolution
layers, in which each convolution layer is accompanied by a BN layer and a
ReLU layer (Fig.~\ref{fig:networkmodule}). While in the residual case, when
size of feature maps doubles, \eg~$16$ to $32$, we use $3\times3$ projection
shortcuts instead of identity ones. All the hyperparameters such as weight decay,
momentum and learning rate are identical with~\cite{He2015Residual}. Horizontal
flip is the only data augmentation.

\textbf{Protocol of ImageNet 2012 classification.}
The architectures in this protocol are also with a slight variation. Detailed
architectures can be referred in Table~\ref{table_architecture}. The
hyperparameters are identical with those of CIFAR-10 protocol. $224\times224$
crops are randomly sampled on $256\times256$ images plus horizontal flip and
color augmentation~\cite{Krizhevsky2012ImageNet}. Mean of RGB is subtracted then
scaling with a factor $0.017$ (standard deviation of RGB). The mini-batch size
is $256$.
Only the performances on validation set are reported.
\begin{table}\scriptsize
\begin{center}
\begin{tabular}{c|c|c|c}
\hline
layer name & output size & 34-layer & 101-layer \\
\hline
conv1 & $112\times112$ & \multicolumn{2}{c}{$7\times7$, $64$, stride $2$}\\
\hline
\multirow{2}{*}[-1.25em]{conv2\_x} & \multirow{2}{*}[-1.25em]{$56\times56$} &
\multicolumn{2}{c}{$3\times3$ max pooling, stride $2$}\\
\cline{3-4}
&  & $\left[\begin{array}{c}3\times3, 64\\3\times3, 64\end{array}\right]\times3$ &
$\left[\begin{array}{c}1\times1, 64\\1\times1, 64\\3\times3, 256\end{array}\right]\times3$\\
\hline
conv3\_x & $28\times28$ & $\left[\begin{array}{c}3\times3, 128\\3\times3, 128\end{array}\right]\times4$
&$\left[\begin{array}{c}1\times1, 128\\1\times1, 128\\3\times3, 512\end{array}\right]\times4$\\
\hline
conv4\_x & $14\times14$ & $\left[\begin{array}{c}3\times3, 256\\3\times3, 256\end{array}\right]\times6$
&$\left[\begin{array}{c}1\times1, 256\\1\times1, 256\\3\times3, 1024\end{array}\right]\times23$\\
\hline
conv5\_x & $7\times7$ & $\left[\begin{array}{c}3\times3, 512\\3\times3, 512\end{array}\right]\times3$
&$\left[\begin{array}{c}1\times1, 512\\1\times1, 512\\3\times3, 2048\end{array}\right]\times3$\\
\hline
&$7\times7$ & $3\times3, 1024$&\\
\hline
 & $1\times1$ & \multicolumn{2}{c}{average pool, 1000-d fc, softmax}\\
\hline
\end{tabular}
\end{center}
\caption{Architectures for ImageNet. Downsampling is performed by conv3\_1, conv4\_1,
and conv5\_1 with a stride of $2$.}
\label{table_architecture}
\end{table}

\subsection{Orthonormality Regularization Enhances the Magnitude of Signals}
In this section, we design experiments to show that orthonormality can indeed enhance the
magnitude of propagated signals in deep plain networks through decorrelating learned
weights among different channels. A 44-layer plain network and CIFAR-10 dataset is adopted.

\begin{figure}[t]
\begin{center}
   \includegraphics[width=0.6\linewidth]{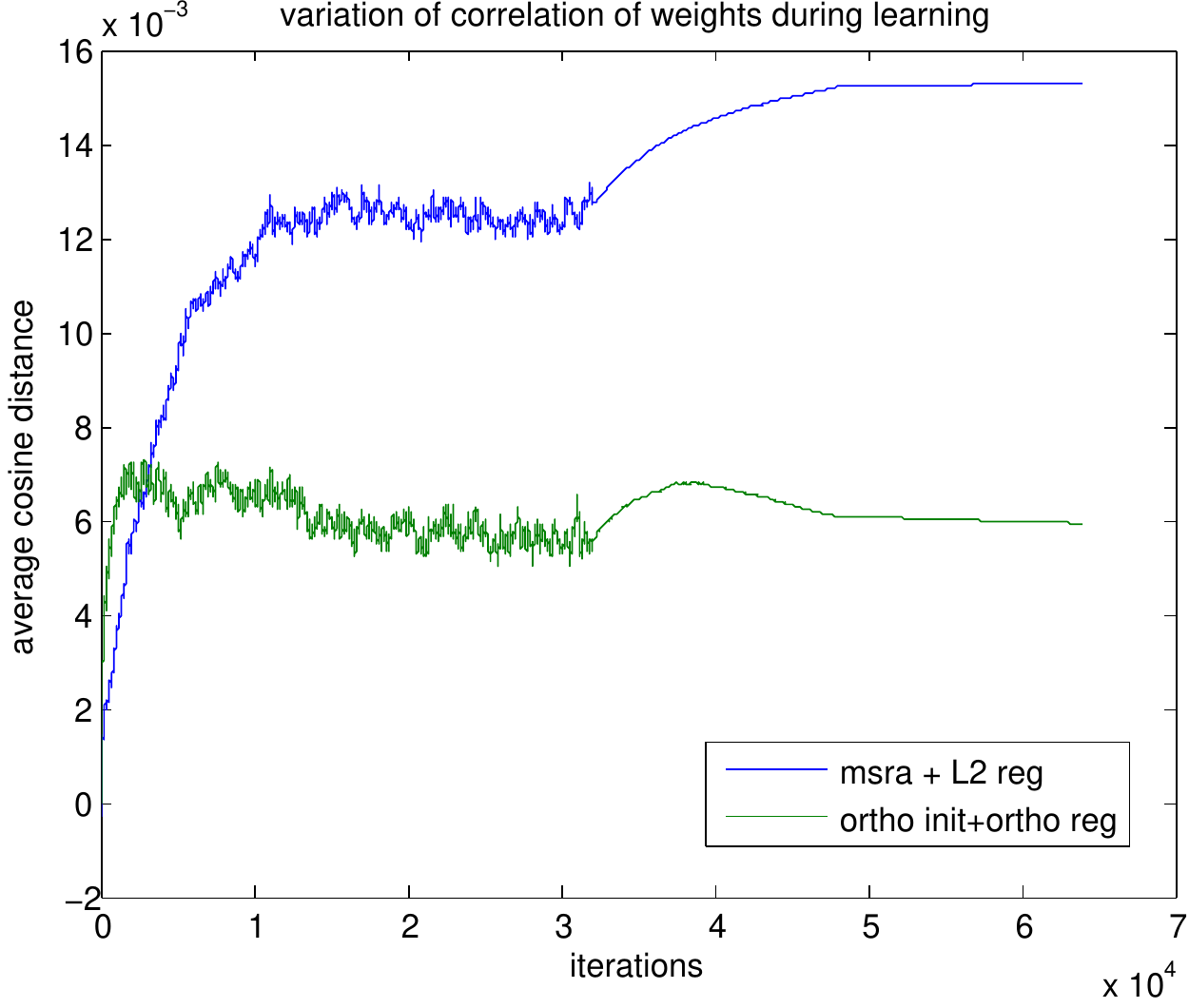}
\end{center}
   \caption{The variation of correlation among weight vectors in 44-layer plain network. The meaning of blue and green line
   can be refereed to the legend. One may be aware of at very first phase the correlation of
   blue line is lower than green one because of we allow negative correlations and ``msra+$L2$ reg"
   method generates negative ones at first few iterations.}
\label{fig:weight_coeff}
\end{figure}

First we make statistics of average correlation among different channels over all the layers
between two types of methods, say ``msra"~\cite{He2015Delving} initialization plus $L2$
regularization (abbr. as ``msra+$L2$ reg") and our proposed orthonormal initialization and
orthonormality regularization (abbr. as ``ortho init+ortho reg").
Cosine distance $D_{cos}(\textbf{x},\textbf{y})$ is considered to compute this value:
\begin{equation}
\bar{s}=\frac{1}{N}\sum_{l=1}^{D}\sum_{i=1}^{f_{out}}\sum_{j=i}^{f_{out}}
D_{cos}(\textbf{v}_{i}^{(l)},\textbf{v}_{j}^{(l)})
\end{equation}
where $\textbf{v}_{i}^{(l)}\in\Re^{f_{in}}$ denotes $i$th kernel of $\textbf{W}_{l}$ in $l$th
layer and $N$ is total computation count.
From Fig.~\ref{fig:weight_coeff} we can see the variation of correlation among weights with
iterations. Under the constraints of orthonormality, correlation of learned weights are forced
into a consistent and relatively lower level (about $6\times10^{-3}$). On the contrary,
``msra+$L2$ reg" cannot prevent from increasing correlation among weights as learning progresses.
Finally, the correlation of ``msra+$L2$ reg" is about $2.5$ times higher than that of
``ortho init+ortho reg", which demonstrates the effectiveness of orthonormality constraints.

\begin{figure}[t]
\begin{center}
   \includegraphics[width=0.7\linewidth]{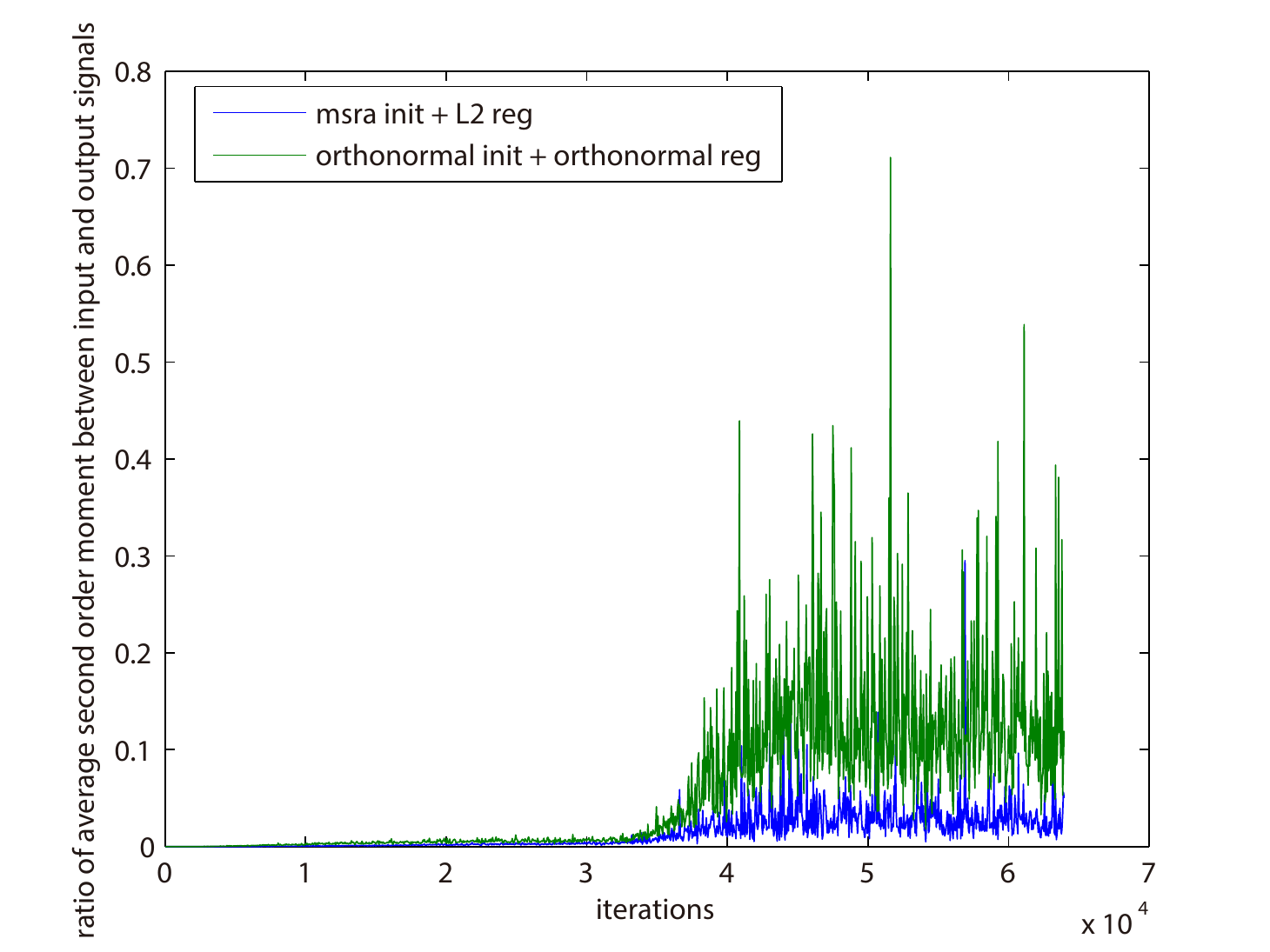}
   \includegraphics[width=0.7\linewidth]{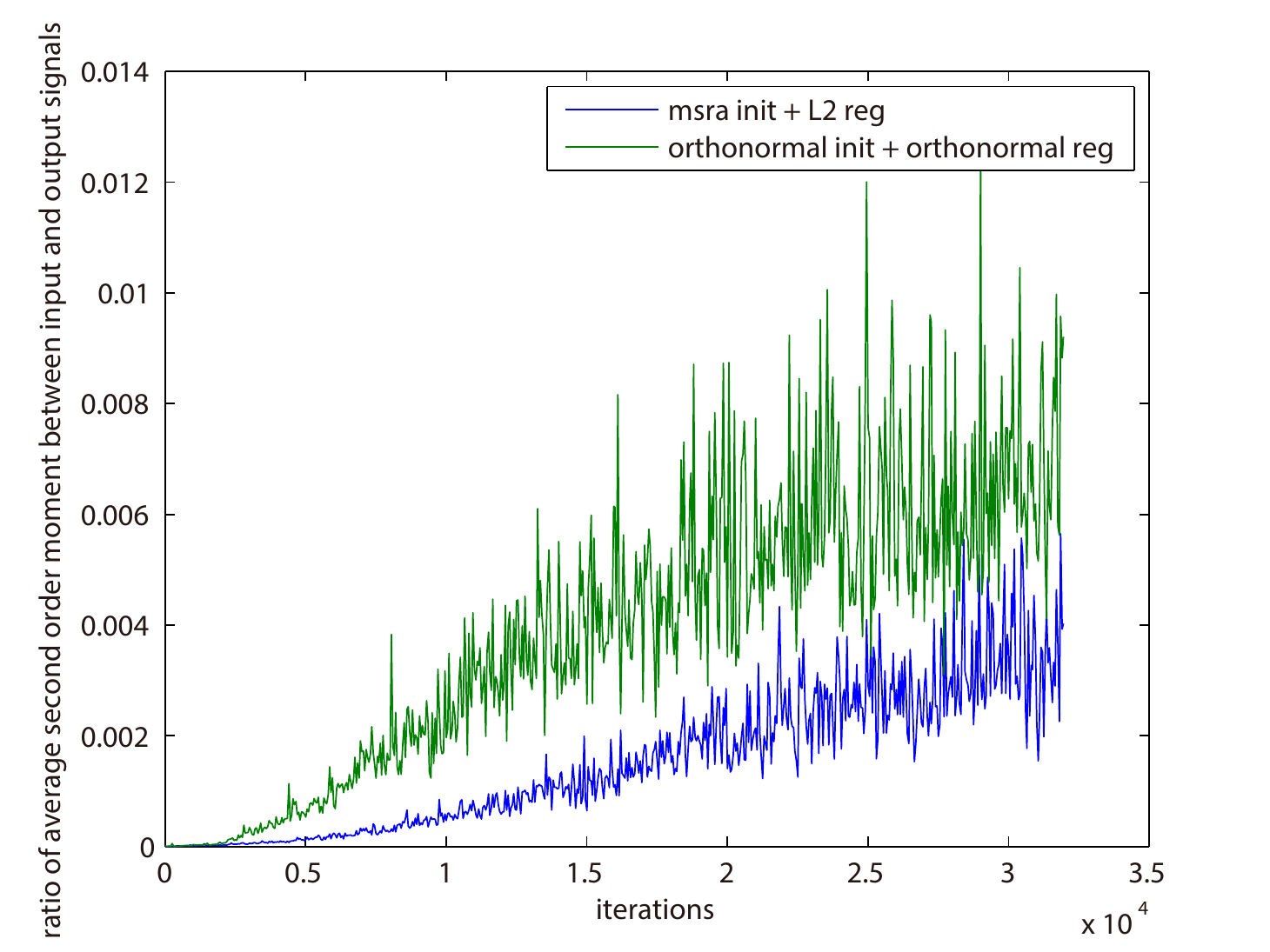}
\end{center}
   \caption{Variation of second order moments ratios of back-propagated errors in 44-layer plain network. Sample interval is $50$ iterations for clearness. Orthonormality regularization can effectively alleviate the vanishing trend of magnitude of signals. \textbf{Top:}~the plot over all iterations. \textbf{Bottom:}~enlarged plot from $5$th iteration to $32000$th iteration (before learning rate is divided by $10$).}
\label{fig:ratio_of_som}
\end{figure}
Next we make statistics of variation of second order moments of back-propagated errors.
Since the empirical risk will convergence as learning progresses, which results in
smaller magnitude of loss value hence unscaled magnitude of error signals, we actually
plot the ratio of second order moment of output error signals (input errors of first
convolution layer) to that of input error signals (input errors of last convolution layer).
Fig.~\ref{fig:ratio_of_som} tells us that in first training phase (when learning rate is relatively
large) the evolution of signal propagation is more insensitive than the in second and third
training phases (when learning rate is small) because of mismatched order of magnitudes between
learning rate and decay coefficient of regularizer ($0.1$ to $0.0001$). However, it shows
the advantage of orthonormal regularization against $L2$ regularization no
matter in which phase, especially in later phases. The magnitude of propagated signals is
enhanced one order of magnitude by orthonormality. It is important to note that we omit the ratios of first five iterations in Fig.~\ref{fig:ratio_of_som} since the disproportional order of magnitude. An interesting phenomenon is that all the magnitude of error signals is vanishing, \eg~ratio is less than $1$, except for the initialization phase, in which the signals are amplified. We think randomness plays the key role for this phenomenon and it also provides evidence that makes us introduce orthonormality beyond initialization in optimizing extremely deep networks.

\subsection{The Rationality of Modulation}
In this section, we present our findings in training deep plain networks and aim to demonstrate modulation is a promising mechanism to train genuinely deep networks.

We find that a 44-layer network can be trained well just with orthonormality but a 110-layer one  incurs seriously divergence, which states the accumulation effect mentioned in Sec.~\ref{sec:BN} by evidence. The proposed modulation is applied to train the 110-layer network and achieves distinct performance improvement against other one-order methods (see Table~\ref{comparison}). The training methodology is a little tricky that we first apply with both orthonormality and modulation at the first $n$ iterations, then the signals are regulated only through orthonormality until it converges. Keeping the magnitude of error signals to be isometric can easily be done by our modulation, but it is observed that this strategy undermines propagation of signals ($81.6\%$ \vs $73.5\%$ on CIFAR-10). So when and how to modulation is an interesting and key research topic to totally solve the degradation problem.

\begin{figure}[t]
\begin{center}
   \includegraphics[width=0.45\linewidth]{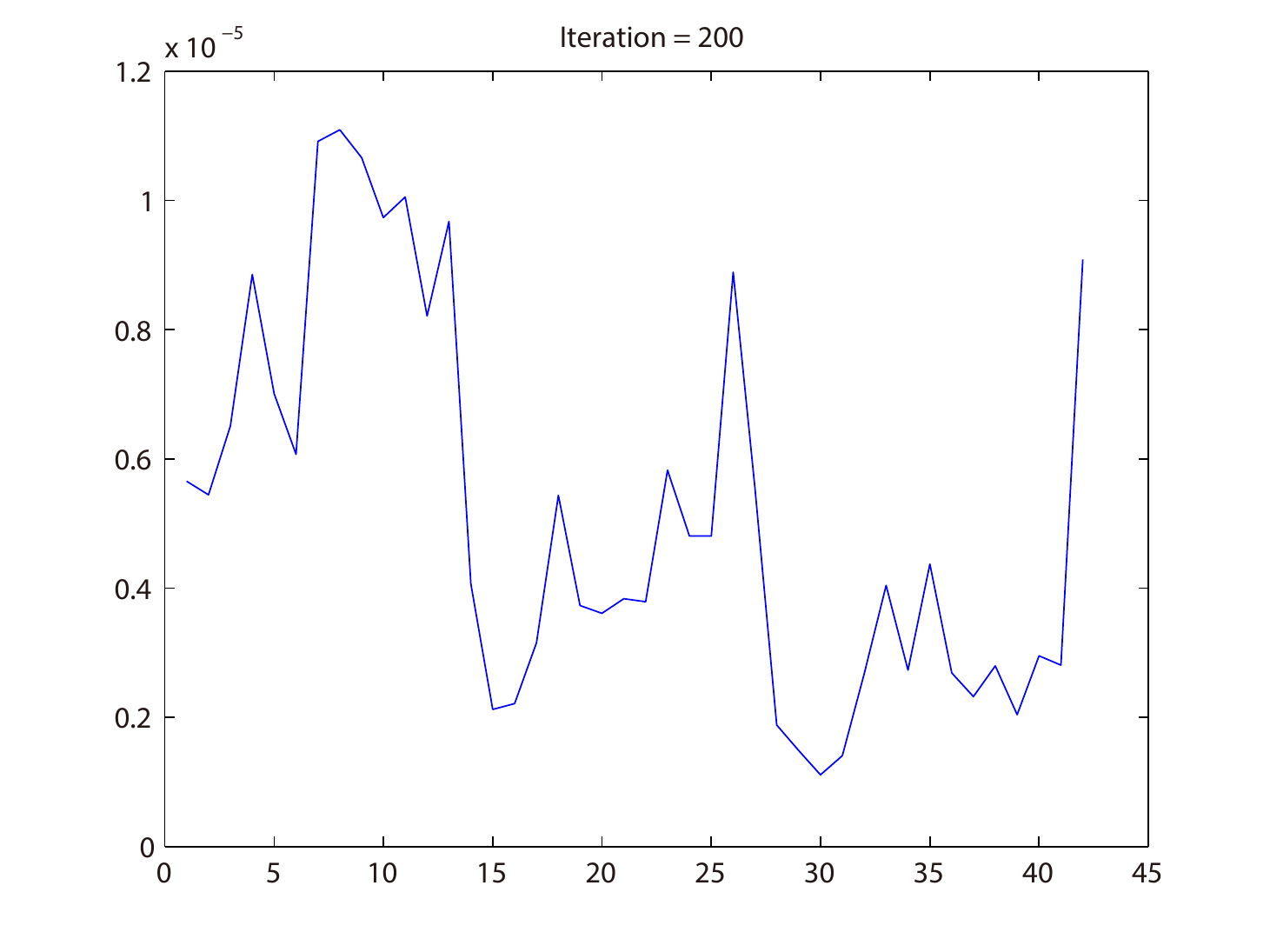}
   \includegraphics[width=0.45\linewidth]{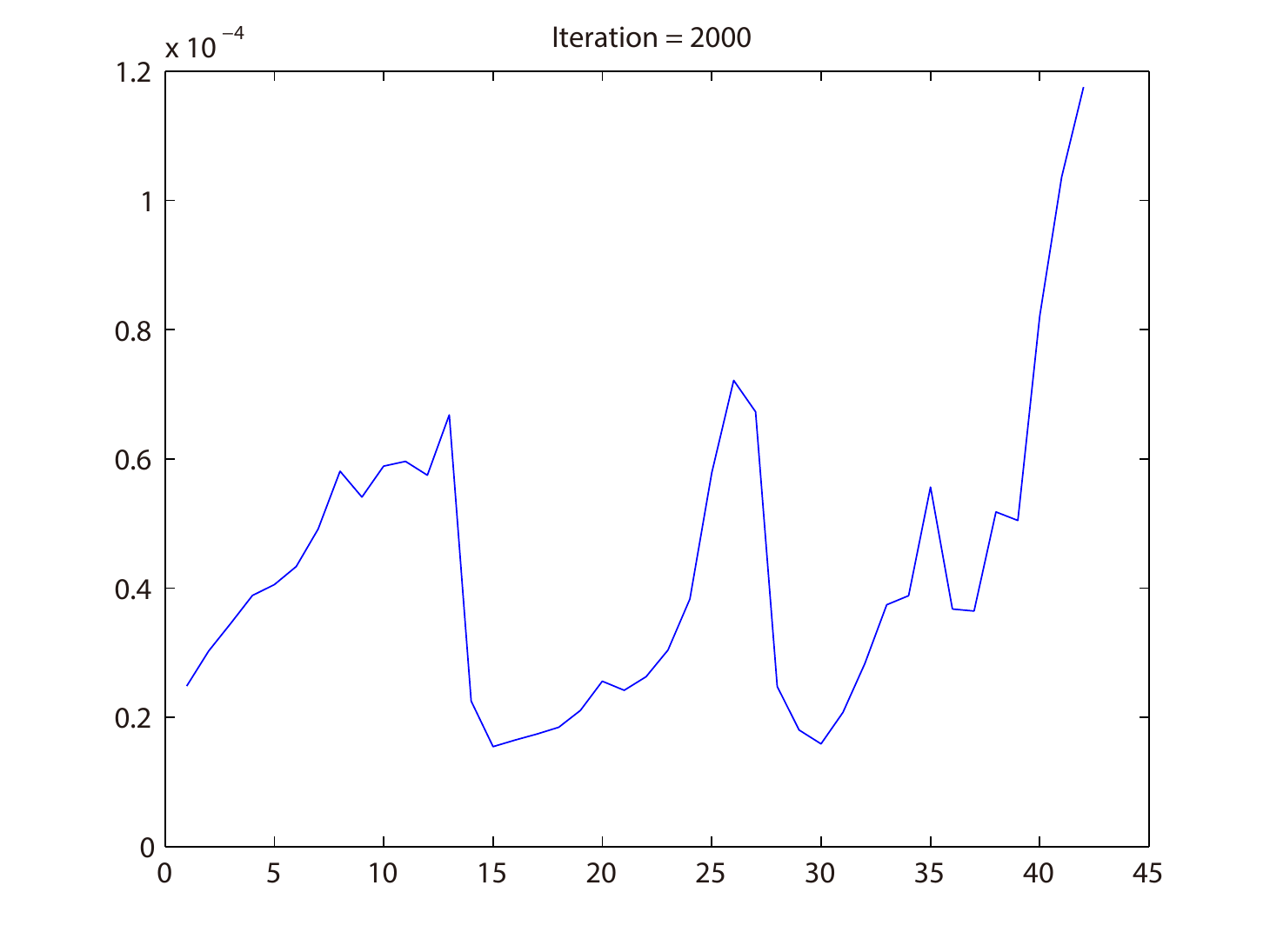}
   \includegraphics[width=0.45\linewidth]{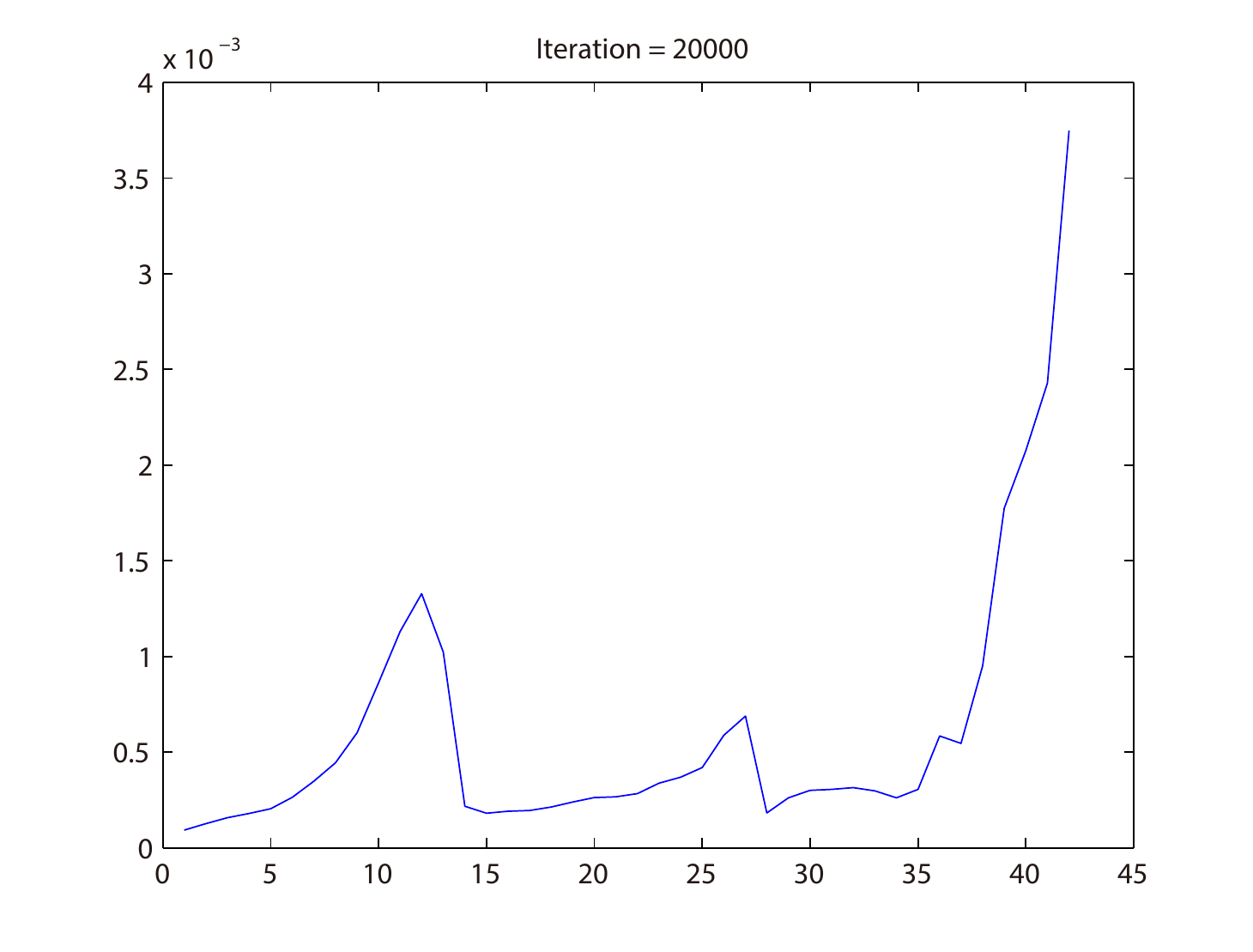}
   \includegraphics[width=0.45\linewidth]{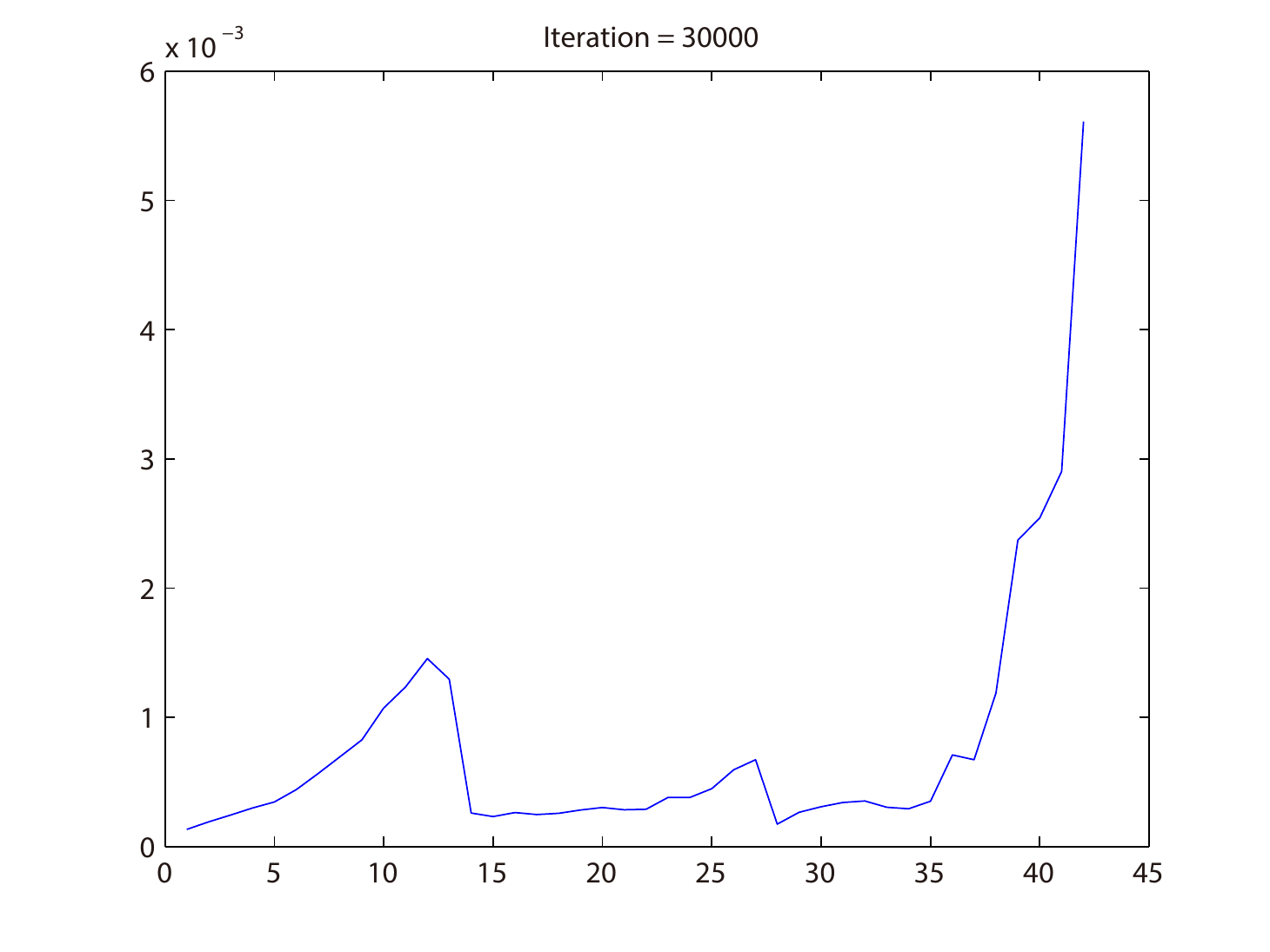}
\end{center}
   \caption{Evolution of backward signal propagation of a 44-layer plain network. $X$ axis denotes layer index and $Y$ axis denotes ratio of second-order moment of current layer to highest layer. We only present $200$th, $2000$th, $20000$th and $30000$th iteration, respectively. About after 2000 to 3000 iterations, the ratio trend to converge to a certain of stable evolution pattern shown in the $20000$th and $30000$th iterations.}
\label{fig:evolution}
\end{figure}

In this paper the value of $n$ is somewhat heuristic, which is derived from our observation to the evolution of ratios of second-order moment of output errors of each layer to the second-order moment of input errors at each iteration of training a 44-layer network. Fig.~\ref{fig:evolution} reveals that it probably exists a potential evolution pattern in training deep networks. Actually we just shrink the degradation gap instead of eliminating it in training genuinely deep networks and one of our future work will focus on the methodology of modulation.
\subsection{Results of Plain and Residual Network Architecture}
To prove our proposed method has advantage against other methods integrated with the idea of adaptivity in training extremely deep plain networks, we compare it with six prevalent one-order methods in this section. We do not compare with second-order methods in consideration of implementation and memory practicality. Table~\ref{comparison} shows the performances\footnote{We should mention that since the particularity of AdaDelta, which is less dependent on the learning rate, for more reasonable comparison, we ignore this hyper-parameter.}. We can see that most methods cannot handle relatively shallow networks well other than SGD and ours and all the methods except for ours cannot even converge in the deeper version. As pointed by~\cite{Sutskever13}, most one-order methods can only be a very effective method for optimizing certain types of deep learning architectures. So next we will focus on making comparison against more general SGD method. We also do not compare our method with other modulation methods, \eg~\cite{Arpit2016Normalization}, because of they will fail convergence at the very first few iteration in such deep architecture
%\footnote{Actually, for NormProp~\cite{Arpit2016Normalization}, the magnitudes of both forward and backward signals explode in our implementation which results in \textsf{NaN}.}.

\begin{figure*}[t]
\begin{center}
    \includegraphics[width=0.33\linewidth]{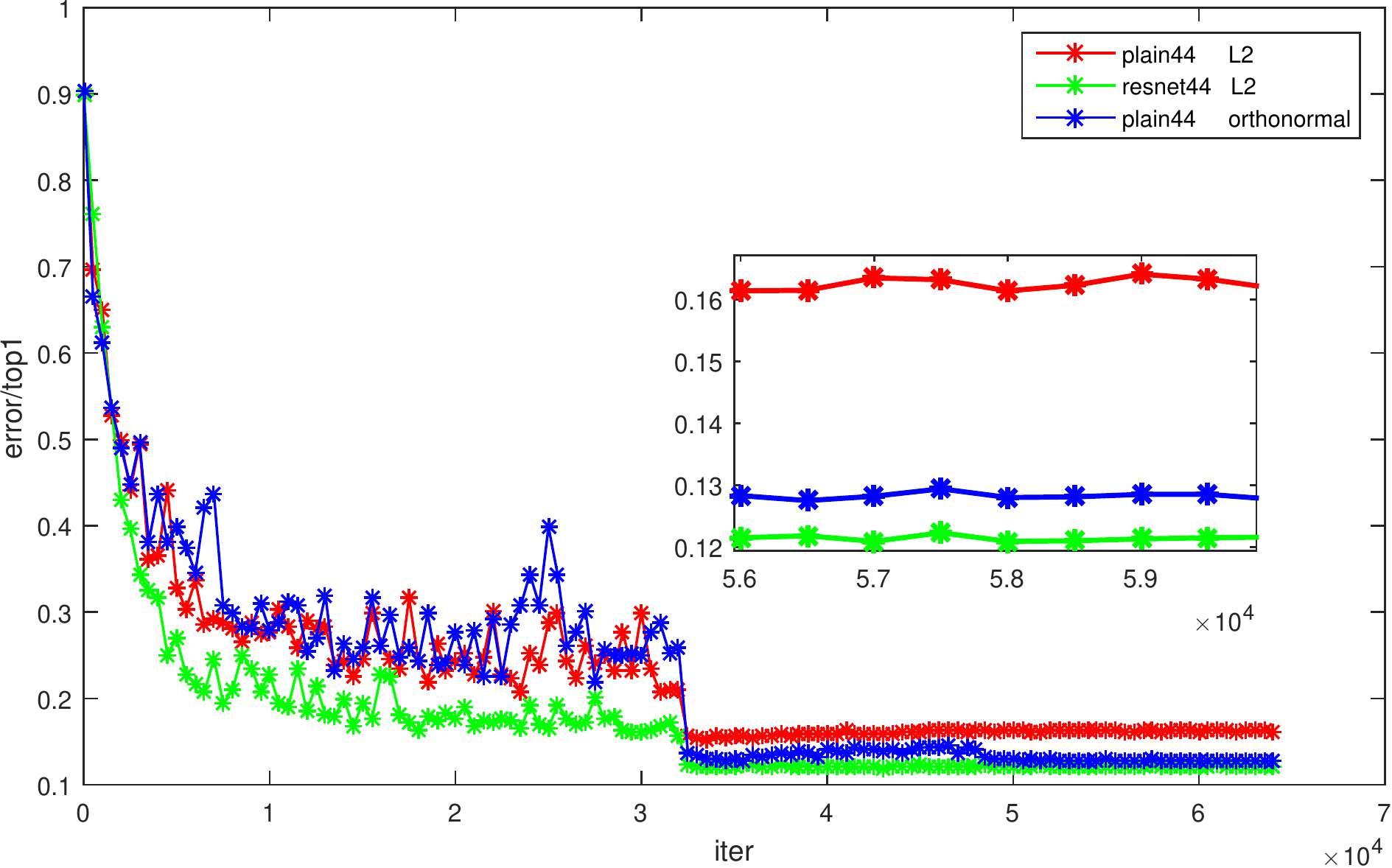}
    \includegraphics[width=0.33\linewidth]{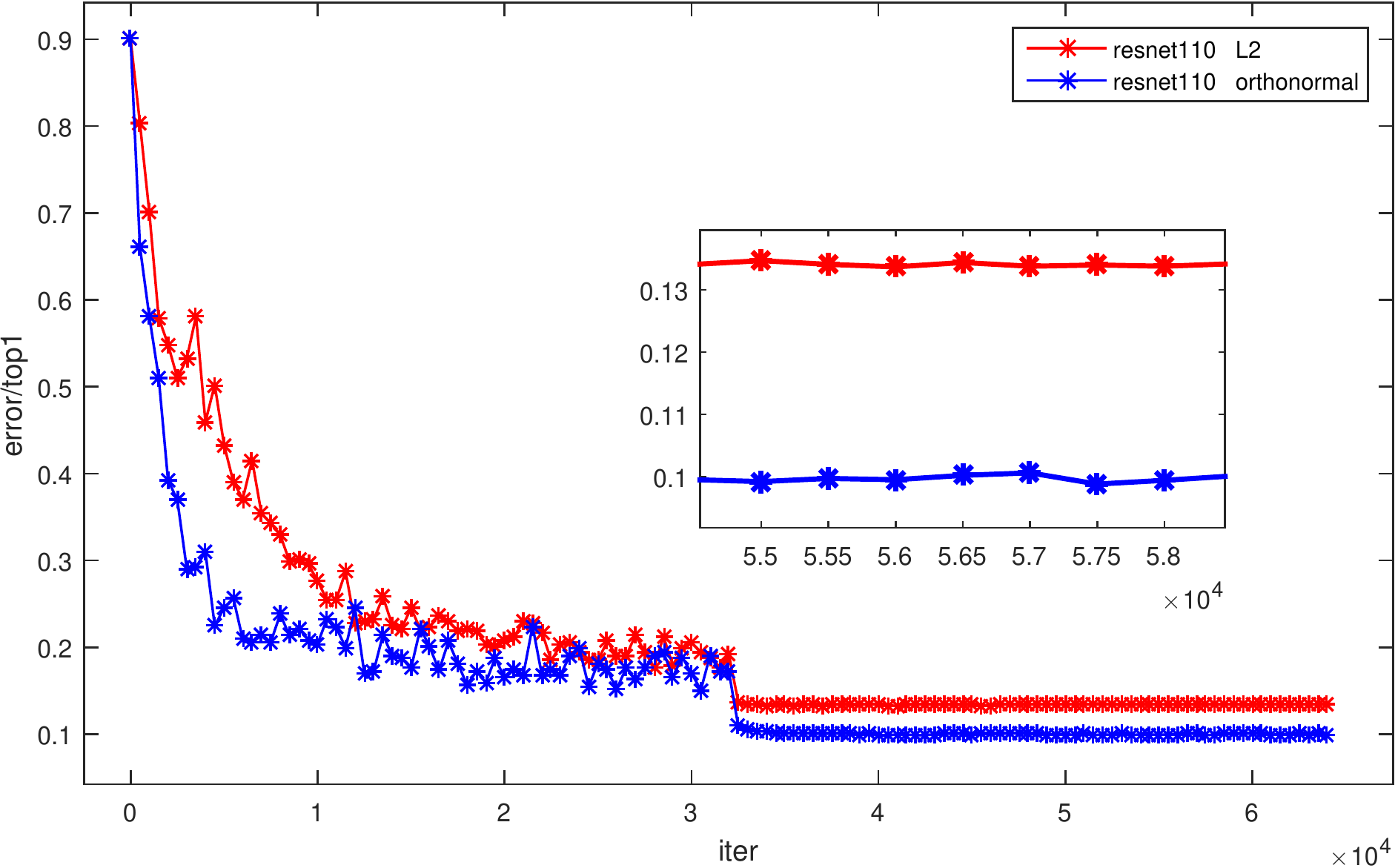}
    \includegraphics[width=0.33\linewidth]{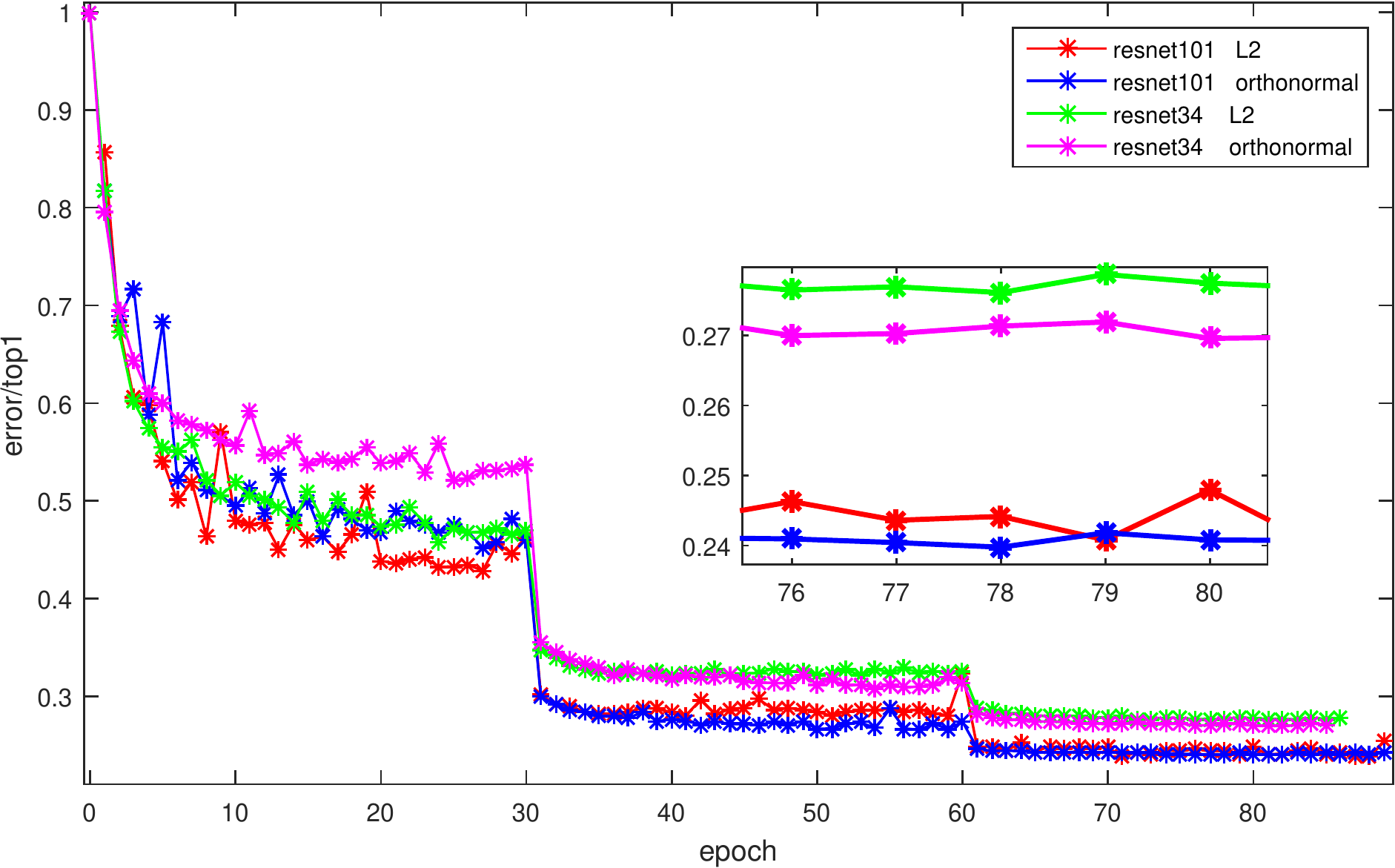}
\end{center}
   \caption{Miscellaneous performance comparisons about plain and residual networks on CIFAR-10 and ImageNet. \textbf{Left:}~Performance comparisons of 44-layer plain network on CIFAR-10. One can see orthonormality boosts plain network to match the performance of residual architecture. \textbf{Middle:}~Performance comparisons of 110-layer ResNet on CIFAR-10. Orthonormality helps convergence thus achieve higher performance. \textbf{Right:}~Performance comparisons of 34-layer ResNet and 101-layer ResNet with different regularization on ImageNet.}
\label{fig:results}
\end{figure*}

Then we compare the performance with different regularizer in identical network architecture (ortho \vs $L2$ of a plain network), and further compare the performance of plain networks with residual networks have similar architectures (plain network with orthonormality \vs residual network with $L2$). Results are shown in Fig.~\ref{fig:results}. We can conclude that our proposed method has distinct advantage in optimizing plain networks and the orthonormality indeed can enhance magnitude of signal which alleviates gradient vanishing in training process.

\begin{figure}[t]
\begin{center}
   \includegraphics[width=0.45\linewidth]{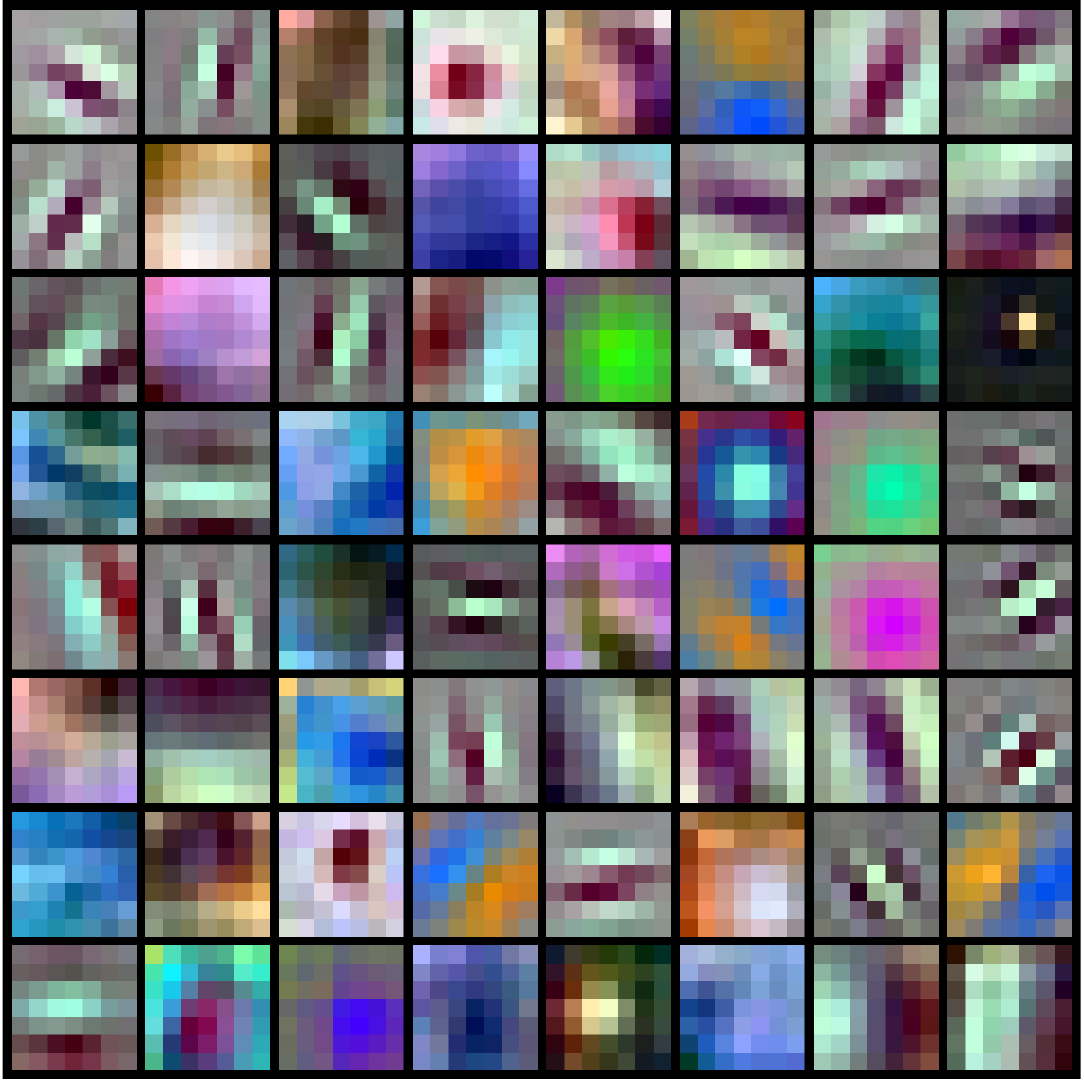}
   \includegraphics[width=0.45\linewidth]{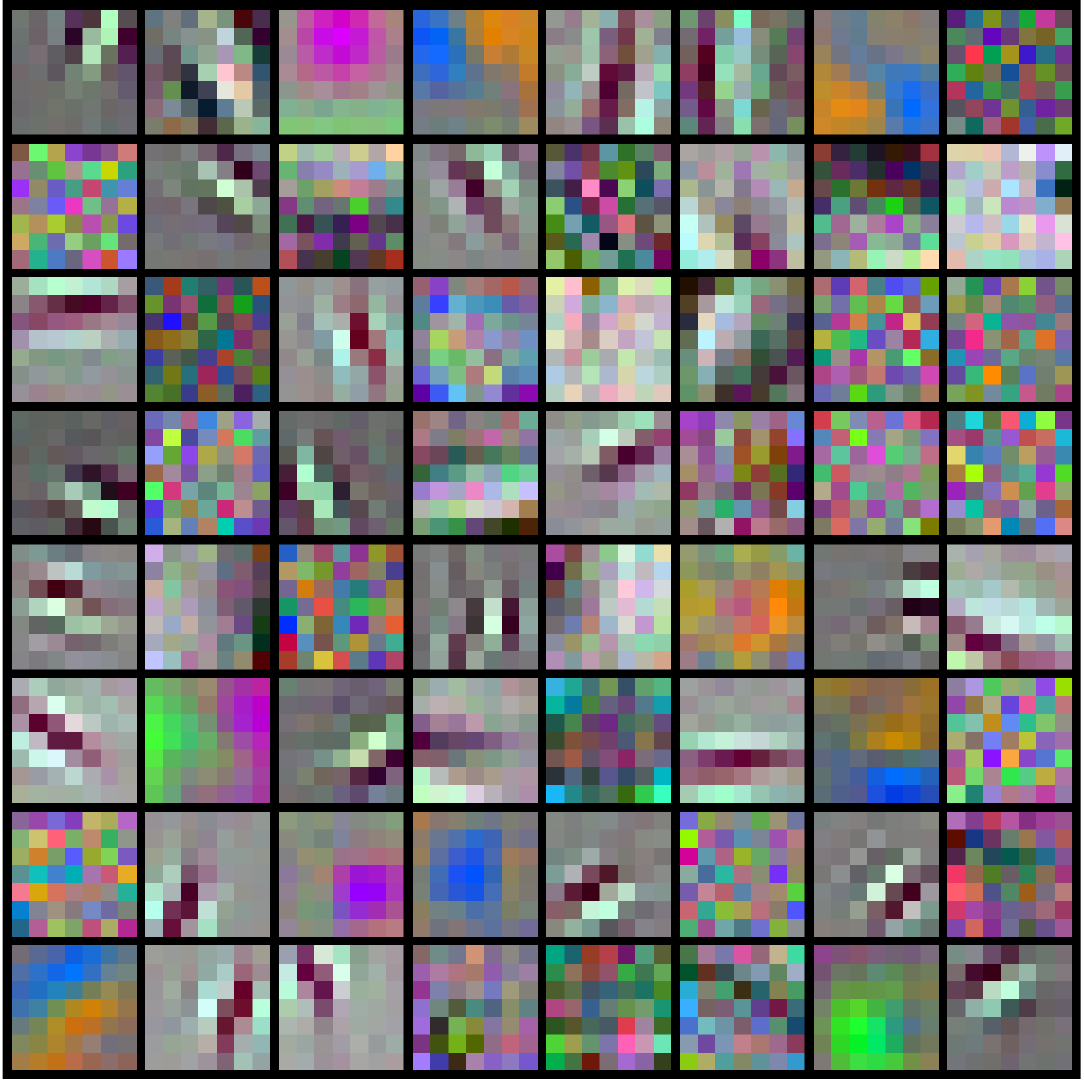}
\end{center}
   \caption{Weights visualization of first convolution layer in 34-layer residual network on ImageNet. \textbf{Left:}~converged weights by $L2$ regularization. \textbf{Right:}~converged weights by orthonormality.}
\label{fig:weights_vis}
\end{figure}

To emphasize that orthonormality can be general to prevalent network architectures and large-scale datasets, we extend the experiments on ImageNet dataset. From Fig.~\ref{fig:results} it shows the decreasing performance boost in ResNet-34 and almost comparative performance in ResNet-110. Compared with architectures on CIFAR-10, they have more channels, \eg~64 \vs 2048, which introduces more redundancies among intra-layer's filter banks. Fig.~\ref{fig:weights_vis} can be used to explain above results, so it probably be difficult for orthonormality to explore in parameter space with so many redundancies. The right sub-figure in Fig.~\ref{fig:weights_vis} shows more noise-like feature maps than the left one, which inspires us to design thinner architectures in the future work.
\begin{table}
\begin{center}
\begin{tabular}{c|c|c|c}
\hline
\multirow{2}{*}{Method} & \multicolumn{3}{c}{Top-1 Accuracy (\%)}\\
\cline{2-4}
& 44-layer & 110-layer & 44-layer*\\
\hline
Nesterov\cite{Nesterov} & 85.0 &10.18&61.9\\
AdaGrad\cite{Duchi11} & 77.86 &30.3&36.1\\
AdaDelta\cite{Zeiler12} & 70.56 &66.48&52.6\\
Adam\cite{Ba15} & 39.85 &10.0&N/A\\
RmsProp\cite{Hinton2012} & 10.0 &10.0&N/A\\
SGD & 84.14 & 11.83&65.2\\
Ours & \textbf{88.42} & \textbf{81.6}&\textbf{70.0}\\
\hline
\end{tabular}
\end{center}
\caption{Performance comparison on CIFAR-10 and ImageNet of different optimization methods. Plain 44-layer and 110-layer networks are trained with these methods. All the common hyperparameters are identical and specific ones are default~(except for AdaDelta). $L2$ regularizer is applied for all the methods except for ours. N/A demonstrates the corresponding method cannot convergence at all and ``*" means the methods are tested on ImageNet.}
\label{comparison}
\end{table}

\section{Discussion and Conclusion}
Recently we find that~\cite{Ozay2016Opt} has proposed similar ideas. They unify three
types of kernel normalization methods into a geometric framework called kernel
submanifolds, in which sphere, oblique and compact Stiefel manifolds (orthonormal kernels)
are considered.
The differences exists in three aspects: 1) The intrinsic explanations about the
performance improvement is different, of which they mainly focus on regularization of models
with data augmentation and learning of models endowed with geometric invariants; 2) The
orthogonalization is different, of which they orthogonalize convolutional kernels within
a channel while we do this among channel; 3) As the second statement tells, we believe that
their proposed method still cannot handle the extremely deep plain networks. Besides, all
the details and key steps to implement their methods are ambiguous that prevents from
understanding and verifying it further.

Intrinsically, one can regard our proposed modulation as assigning each parametric layer an individual and adaptive learning rate. This kind of modulation can be more practical than local methods, \eg~second-order methods, while be more flexible than global ones, \eg~SGD. Besides, if we can approach some strategies to compensate the evanescent orthonomality as learning progresses, we believe that training a genuinely deep network will be available.

We propose a simple and direct method to train extremely deep plain networks with orthonormality and modulation. Furthermore, orthonormality reveals its generalization capability which can be applied in residual networks. Great performance boost is observed in experiments. However, the degradation problem is still not totally solved, which may be on condition understanding more comprehensively about the insights of signal modulation, reparametrization and novel constraints, \etc. We hope our work will encourage more attentions on this problem.

%-------------------------------------------------------------------------

{\small
\bibliographystyle{ieee}
\bibliography{egbib}
}

\section{Quasi-isometry inference with Batch Normalization}

%Please follow the steps outlined below when submitting your manuscript to
%the IEEE Computer Society Press.  This style guide now has several
%important modifications (for example, you are no longer warned against the
%use of sticky tape to attach your artwork to the paper), so all authors
%should read this new version.
For batch normalization (BN) layer, its Jacobian, denoted as $\textbf{J}$, is not only related with components of activations ($d$ components in total), but also with samples in one mini-batch (size of $m$).

Let $x_{j}^{(k)}$ and $y_{i}^{(k)}$ be $k$th component of $j$th input sample and $i$th output sample respectively and given the independence between different components, $\frac{\partial y_{i}^{(k)}}{\partial x_{j}^{(k)}}$ is one of $m^{2}d$ nonzero entries of $\textbf{J}$. In fact, $\textbf{J}$ is a tensor but we can express it as a blocked matrix:
\begin{equation}
\textbf{J}=\left[
\begin{array}{cccc}
\textbf{D}_{11} & \textbf{D}_{12} & \cdots & \textbf{D}_{1m}\\
\textbf{D}_{21} & \textbf{D}_{22} & \cdots & \textbf{D}_{2m}\\
\vdots & \vdots & \ddots & \vdots\\
\textbf{D}_{m1} & \textbf{D}_{m2} & \cdots & \textbf{D}_{mm}
\end{array}
\right]
\label{eq_J}
\end{equation}
where each $\textbf{D}_{ij}$ is a $d\times d$ diagonal matrix:
\begin{equation}
\textbf{D}_{ij}=\left[
\begin{array}{cccc}
\frac{\partial y_{i}^{(1)}}{\partial x_{j}^{(1)}} &  &  & \\
 & \frac{\partial y_{i}^{(2)}}{\partial x_{j}^{(2)}} & & \\
&  & \ddots & \\
 & &  & \frac{\partial y_{i}^{(d)}}{\partial x_{j}^{(d)}}
\end{array}
\right]
\end{equation}

Since BN is a component-wise rather than sample-wise transformation, we prefer to analyse a variant of Eq.~\ref{eq_J} instead of $\textbf{D}_{ij}$. Note that by elementary matrix transformation, the $m^{2}$ $d\times d$ matrices can be converted into $d$ $m\times m$ matrices:
\begin{equation}
\textbf{J}=\left[
\begin{array}{cccc}
\textbf{J}_{11} & \textbf{0} & \cdots & \textbf{0}\\
\textbf{0} & \textbf{J}_{22} & \cdots & \textbf{0}\\
\vdots & \vdots & \ddots & \vdots\\
\textbf{0} & \textbf{0} & \cdots & \textbf{J}_{dd}
\end{array}
\right]
\end{equation}
and the entries of each $\textbf{J}_{kk}$ is
\begin{equation}
\frac{\partial y_{j}}{\partial x_{i}}=\rho\left[\Delta(i=j)-\frac{1+\hat{x}_{i}\hat{x}_{j}}{m}\right]
\label{eq_Jkk}
\end{equation}
The notations of $\rho$, $\Delta(\cdot)$ and $\hat{x}_{k}$ have been explained in our main paper and here we omit the component index $k$ for clarity. Base on the observation of Eq.~\ref{eq_Jkk}, we separate the numerator of latter part and denote it as $U_{ij}=1+\hat{x}_{i}\hat{x}_{j}$.

Let $\hat{\textbf{x}} = (\hat{x}_{1},\hat{x}_{2},...,\hat{x}_{m})^T$, $\textbf{e} = (1,1,..1)^T$, we have
\begin{equation}
\textbf{U} = \textbf{e}\textbf{e}^T + \hat{\textbf{x}}\hat{\textbf{x}}^{T}
\end{equation}
and
\begin{equation}
\textbf{J}_{kk} = \rho(\textbf{I} - \frac{1}{m}\textbf{U})
\label{eq_Jkkm}
\end{equation}

Recall that for any column vector $\textbf{v}$, $\verb"rank"(\textbf{v}\textbf{v}^{T}) = 1$. According to the subadditivity of matrix rank~\cite{Banerjee2014Linear}, it implies that
\begin{equation}
\begin{aligned}
\verb"rank"(\textbf{U}) = \verb"rank"(\textbf{e}\textbf{e}^{T}+\hat{\textbf{x}}\hat{\textbf{x}}^{T})\leq\\ \verb"rank"(\textbf{e}\textbf{e}^{T}) + \verb"rank"(\hat{\textbf{x}}\hat{\textbf{x}}^{T})=2
\end{aligned}
\label{eq_rankU}
\end{equation}

Eq.~\ref{eq_rankU} tells us that $\textbf{U}$ actually only has two nonzero eigenvalues, say $\lambda_1$ and $\lambda_2$, and we can formulate $\textbf{U}$ as follow:
\begin{equation}
\textbf{U}=\textbf{P}^{T}
\left[
\begin{array}{ccccc}
\lambda_{1} &  &  &  & \\
 & \lambda_{2} &  &  & \\
 &  & 0 &  & \\
 & & & \ddots & \\
 &  &  &  & 0
\end{array}
\right]
\textbf{P}
\end{equation}
combined with Eq.~\ref{eq_Jkkm}, finally we get the equation of $\textbf{J}_{kk}$ from the eigenvalue decomposition view, which is
\begin{equation}
\textbf{J}=\textbf{P}^{T}
\rho\left[
\begin{array}{ccccc}
1-\frac{\lambda_{1}}{m} &  &  &  & \\
 & 1-\frac{\lambda_{2}}{m} &  &  & \\
 &  & 1 &  & \\
 & &  & \ddots & \\
 &  &  & & 1
\end{array}
\right]
\textbf{P}
\end{equation}

To show that $\textbf{J}_{kk}$ probably is not full rank, we formulate the relationship between $\textbf{U}^{2}$ and $\textbf{U}$
\begin{equation}
%\begin{flalign*}
%\begin{split}
\begin{aligned}
\textbf{U}^{2} = (\textbf{e}\textbf{e}^{T}+\hat{\textbf{x}}\hat{\textbf{x}}^{T})
(\textbf{e}\textbf{e}^{T}+\hat{\textbf{x}}\hat{\textbf{x}}^{T}) = \textbf{e}\textbf{e}^{T}\textbf{e}\textbf{e}^{T} + \textbf{e}\textbf{e}^{T}\hat{\textbf{x}}\hat{\textbf{x}}^{T} \\+ \hat{\textbf{x}}\hat{\textbf{x}}^{T}\textbf{e}\textbf{e}^{T} + \hat{\textbf{x}}\hat{\textbf{x}}^{T}\hat{\textbf{x}}\hat{\textbf{x}}^{T} = m\textbf{e}\textbf{e}^{T} + (\sum_{i=1}^m\hat{x}_i)\textbf{e}\hat{\textbf{x}}^T\\ + (\sum_{i=1}^m\hat{x}_i)\hat{\textbf{x}}\textbf{e}^T + (\sum_{i=1}^m\hat{x}_i^2)\hat{\textbf{x}}\hat{\textbf{x}}^{T} \\= m\textbf{U} + (\sum_{i=1}^m\hat{x}_i)\textbf{e}\hat{\textbf{x}}^T + (\sum_{i=1}^m\hat{x}_i)\hat{\textbf{x}}\textbf{e}^T + (\sum_{i=1}^m\hat{x}_i^2-m)\hat{\textbf{x}}\hat{\textbf{x}}^{T}
%\end{split}
%\end{flalign*}
\end{aligned}
\label{eq_U2}
\end{equation}

Note that $\hat{x}_i \thicksim N(0,1)$, so we can regard the one-order and second-order accumulated items in Eq.~\ref{eq_U2} as approximately equaling the corresponding one-order and second-order statistical moments for relatively large mini-batch, from which we get $\textbf{U}^{2} \approx m\textbf{U}$.

The relationship implies that $\lambda_{1}^{2}\approx m\lambda_{1}$ and $\lambda_{2}^{2}\approx m\lambda_{2}$. Since $\lambda_{1}$ and $\lambda_{2}$ cannot be zeros, it concludes that $\lambda_{1}\approx\lambda_{2}\approx m$ therefor $1-\frac{\lambda_{1}}{m}\approx 0$ and $1-\frac{\lambda_{2}}{m}\approx 0$ if batch size is sufficient in a statistical sense.
\end{document}